\documentclass[letterpaper, 10pt, conference]{style/ieeeconf} 
\makeatletter
\let\NAT@parse\undefined
\makeatother

\usepackage{dblfloatfix}
\usepackage[numbers, sectionbib, sort]{natbib}
\usepackage{bm}
\usepackage{gensymb}
\usepackage{xcolor}
\usepackage{graphicx}
\usepackage{amsmath}
\usepackage{amssymb}
\usepackage{subcaption}
\usepackage{amsfonts}
\usepackage{siunitx}
\usepackage{booktabs}
\usepackage{makecell}
\usepackage{multirow}
\usepackage{upgreek}
\usepackage[font=small]{caption}
\usepackage[export]{adjustbox}
\usepackage{tikz}
\usepackage{tabularx}
\usepackage{sidecap} \sidecaptionvpos{figure}{c}
\usepackage[hidelinks]{hyperref}
\usepackage[nameinlink, capitalize]{cleveref}
\usepackage[printonlyused,withpage,nolist,nohyperlinks]{acronym}

\Crefname{section}{Sec.}{Sec.}
\Crefname{equation}{Eq.}{Eq.}

\usepackage{soul}

\newcommand{\etal}{\textit{et al}.}
\newcommand{\ie}{\textit{i}.\textit{e}.}
\newcommand{\eg}{\textit{e}.\textit{g}.}

\IEEEoverridecommandlockouts

\title{\Large \bf GO-VMP: Global Optimization for View Motion Planning in Fruit Mapping}

\author{Allen Isaac Jose$^\star$ \and Sicong Pan$^\star$ \and Tobias Zaenker \and Rohit Menon \and Sebastian Houben \and Maren Bennewitz %
\thanks{$^\star$These authors contributed equally to this work.}
\thanks{A. Isaac Jose and S. Houben are with the Bonn-Rhein-Sieg University of Applied Sciences, Germany. A. Isaac Jose is additionally with the Humanoid Robots Lab at the University of Bonn. S. Houben is additionally with the Fraunhofer Institute for Intelligent Analysis and Information Systems. S. Pan, T. Zaenker, R. Menon, and M. Bennewitz are with the Humanoid Robots Lab at the University of Bonn, the Lamarr Institute for Machine Learning and Artificial Intelligence, and the Center for Robotics, University of Bonn, Germany.
This work has partially been funded by the Deutsche Forschungsgemeinschaft (DFG, German Research Foundation) under grant 459376902 – AID4Crops, under Germany’s Excellence Strategy, EXC-2070 – 390732324 – PhenoRob, and by the BMBF within the Robotics Institute Germany, grant No. 16ME0999.
}
}

\begin{document}

\maketitle
\thispagestyle{empty}
\pagestyle{empty}
\begin{abstract}
Automating labor-intensive tasks such as crop monitoring with robots is essential for enhancing production and conserving resources.
However, autonomously monitoring horticulture crops remains challenging due to their complex structures, which often result in fruit occlusions.
Existing view planning methods attempt to reduce occlusions but either struggle to achieve adequate coverage or incur high robot motion costs. 
We introduce a global optimization approach for view motion planning that aims to minimize robot motion costs while maximizing fruit coverage.
To this end, we leverage coverage constraints derived from the set covering problem~(SCP) within a shortest Hamiltonian path problem~(SHPP) formulation. 
While both SCP and SHPP are well-established, their tailored integration enables a unified framework that computes a global view path with minimized motion while ensuring full coverage of selected targets.
Given the NP-hard nature of the problem, we employ a region-prior-based selection of coverage targets and a sparse graph structure to achieve effective optimization outcomes within a limited time.
Experiments in simulation demonstrate that our method detects more fruits, enhances surface coverage, and achieves higher volume accuracy than the motion-efficient baseline with a moderate increase in motion cost, while significantly reducing motion costs compared to the coverage-focused baseline.
Real-world experiments further confirm the practical applicability of our approach.
\end{abstract}

\section{Introduction} \label{S:introduction}

\begin{figure}[!t]
\centering
\includegraphics[width=0.98\columnwidth]{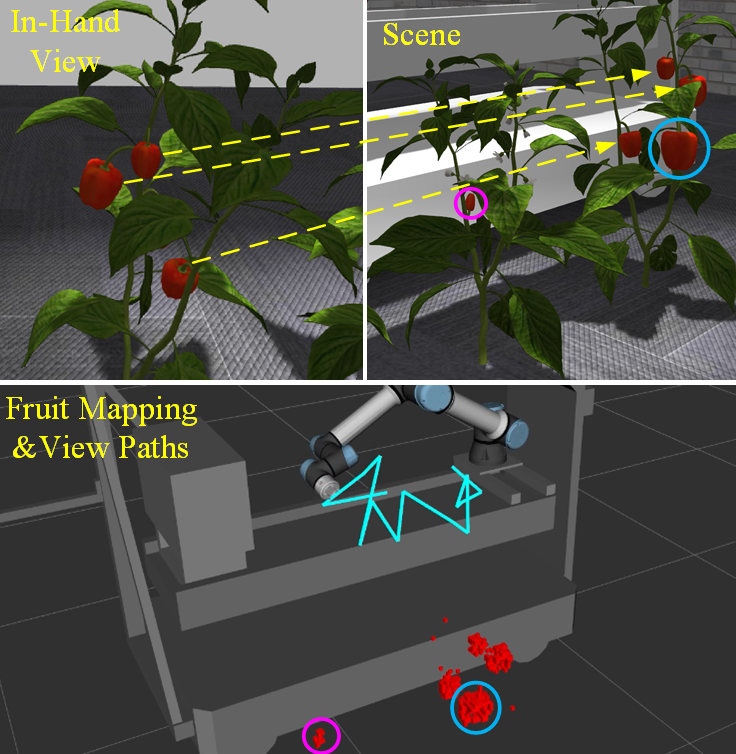}
\caption{
Results of our globally optimized view motion planning~(GO-VMP) for fruit mapping.
One fruit is occluded in the current robot’s in-hand view (blue circle), while another fruit on a different plant still needs to be explored (magenta circle).
Our system successfully maps additional fruits using optimized views and their connecting paths (cyan lines), which are designed to minimize motion cost while maximizing coverage of all fruits.
}
\label{fig_cover}
\vspace{-0.5cm}
\end{figure}

Monitoring crops and understanding their phenology are essential for optimizing yields and diagnosing anomalies. 
Horticulture crops such as sweet peppers and tomatoes are being increasingly cultivated in controlled environments such as glasshouses, where crops are grown along rows. 
In spite of the more structured environment, manual monitoring of produce yield is challenging due to its time-consuming and labor-intensive nature.
To address this, automated mobile platforms are deployed to traverse crop rows and map fruits with an attached robotic arm equipped with RGB-D sensors~\citep{zaenker2021viewpoint,menon2023iros,zaenker2023iros,burusa2024attention,burusa2024semantics}.
However, online fruit mapping in horticulture crops is inherently challenging due to occlusions from leaves, other fruits, and the plants' complex structures, often resulting in insufficient captured fruits in the generated maps with fixed sensor viewpoints~\citep{hemming2014fruit,sa2017peduncle}.
Consequently, effective view planning is required to reduce occlusions by optimizing the robotic arm's trajectory within a designated mission time budget.

Next-best-view~(NBV) planning methods leverage the current state of the environment to identify the most informative viewpoints for data acquisition~\citep{zaenker2021viewpoint,marangoz2022case,menon2023iros,burusa2024gradient}. 
In agricultural settings, candidate viewpoints are typically generated by sampling within the robotic arm’s reachable workspace to focus on regions of interest~(ROIs).
Each candidate view is quantitatively assessed based on its expected information gain.
Although these methods can achieve high fruit coverage, their reliance on greedy next-step path planning lacks a long-horizon perspective, often resulting in suboptimal paths with increased motion costs. 
This is undesirable in practical production, as iLowt requires additional electrical resources and raises operational risks.

In contrast, the view motion planner~(VMP)~\citep{zaenker2023iros} constructs a view motion graph for greedy multi-step planning.
In this graph, vertices represent sampled views, while the edges' weights are determined by robot motion costs.
A sequence of views and connecting paths is then searched on the graph by maximizing the accumulated utility of motion-weighted information gain, resulting in more efficient robot trajectories.
However, this method employs a greedy best-first path search, which does not guarantee full coverage of targets or the global shortest path.
Under a limited mission time budget, its fruit coverage falls short compared to shape prior-based methods such as NBV-SC~\citep{menon2023iros}.

To address these limitations, we propose a global optimization approach that builds upon the concept of the view motion graph~\citep{zaenker2023iros}.
Instead of using greedy search, our method leverages the shortest Hamiltonian path problem~(SHPP) to compute a global view path that minimizes robot motion cost.
However, the original SHPP requires traversing all views on the graph, which is not desirable for our application.
We integrate coverage constraints from the set covering problem~(SCP) into the SHPP formulation to select a subset of views that achieves full coverage of the selected targets.
Since both SCP and SHPP are \mbox{NP-hard}, we sparsify the graph and employ a region-prior-based selection strategy for coverage targets.
Our ablation experiments demonstrate that this approach achieves effective optimization outcomes in a limited time.
Fig.~\ref{fig_cover} showcases the fruit mapping result of our system.

Compared to motion-efficient VMP~\citep{zaenker2023iros}, our method achieves significantly enhanced fruit mapping with a moderate increase in motion cost.
Compared to coverage-focused NBV-SC~\citep{menon2023iros}, our approach significantly reduces motion cost.
Our main contributions are summarized as follows:
\begin{itemize} 
    \item We propose a global optimization formulation that integrates SCP and SHPP for view motion planning in fruit mapping, replacing the conventional greedy search. 
    \item We introduce region-prior-based targets and a sparse graph to achieve an effective sequence of views and connecting paths with a limited optimization time. 
    \item Our globally optimized view motion planner (GO-VMP) achieves a superior balance between fruit mapping performance and robot motion cost.
\end{itemize} 
For reproducibility, our implementation can be accessed at \url{https://github.com/AllenIsaacJose/GO-VMP}.

\section{Related Work} \label{S:related_work}

\subsection{View Planning in Autonomous Agriculture}

An increasing number of studies are exploring the use of robots to enhance autonomy in precision agriculture, including autonomous platforms~\citep{smitt2021pathobot}, selective harvesting~\citep{lenz2024hortibot}, interactive manipulation~\citep{yao2024safe}, and more.
Among these, the most relevant literature pertains to active perception algorithms aimed at reducing occlusions based on view planning. 

Most methods follow the NBV paradigm~\cite{zaenker2021viewpoint,menon2023iros,burusa2024attention,burusa2024semantics,burusa2024gradient,la2024enhancing,ci2025ssl,li2025enhanced}, which maximizes current information gain and iteratively map the crops.
Recent approaches have demonstrated that incorporating shape priors can greatly improve fruit mapping performance~\citep{marangoz2022case,magistri2022contrastive,pan2023panoptic}, which are later integrated into view planning.
For example, Wu~\etal~\citep{wu2019plant} propose a shape prior that predicts a complete plant structure from a partial point cloud to update the occupancy map and guide the selection of NBVs.
Similarly, Menon~\etal~\citep{menon2023iros} leverage fused information from partially visible fruits to fit a superellipsoid as a shape prior, yielding higher information gain by targeting surface regions with more unknown voxels that may contain potential ROIs.
By utilizing these priors, computationally intensive ray casting is avoided, thereby saving planning time.
These NBV works rely on greedy next-step path planning, which can result in longer paths.

Zaenker~\etal~\citep{zaenker2023iros} propose VMP to address this limitation of NBV methods by generating a sequence of views with paths using their view motion graph.
The motion costs are calculated in terms of joint-space distance using MoveIt~\citep{chitta2016ros}, which offers greater accuracy than Euclidean motion costs, especially considering that two nearby points in Euclidean space may be distant in joint space due to joint constraints. 
Look-at voxels are sampled around the frontier voxels between known and unknown areas, and candidate views are sampled uniformly within the workspace oriented in the direction of these look-at voxels. 
The generated path is executed until a fixed time elapses or a collision is detected, at which point the graph is updated and the path search is repeated.
However, it relies on a greedy best-first path search, which does not guarantee full coverage of the target voxels or the global shortest path.
In this work, we integrate the constraints from SCP into the SHPP framework with a sparse graph and a region-based prior to jointly optimize view selection and global path planning.
Our approach ensures complete coverage of selected targets while minimizing robot motion costs.

\subsection{Set Covering Problem in View Planning}

One of the key ideas behind our method is to leverage the constraints from SCP to ensure full coverage of the selected targets.
In model-based view planning~\citep{peuzin2021survey}, where the environment model is available, the SCP is naturally defined for inspection tasks.
Scott~\etal~\citep{scott2009model} proposed a method to translate model-based view planning into an SCP framework by finding the minimal set of views that fully cover the object surfaces.
Some approaches also incorporate path planning during solving the SCP to reduce inspection time~\citep{jing2018computational}.
From an optimization formulation perspective, the work by Wang~\etal~\citep{wang2007metric} is most similar to our work, as it combines traveling cost with view coverage.
However, it emphasizes sensor range and visibility rather than directly modeling target coverage.
Moreover, all these methods rely on a prior 3D model of the environment, which is unavailable in our fruit monitoring task.
In this scenario, the plant structure is unknown beforehand, as it continuously grows up and evolves over time.

In the field of active unknown object reconstruction, a new trend is emerging, in which deep learning is employed to learn priors that serve as a proxy for the inaccessible 3D model, thereby enabling the integration of the SCP into the view planning process.
Recent methods~\citep{pan2024tro,pan2024iros} employ CNN-based networks or large diffusion models to obtain the minimal set of views.
The SHPP is solved to connect these planned views from the current view, thus generating a global shortest view path.
However, this two-stage paradigm is not directly suitable for fruit mapping for two main reasons:
First, this approach decouples view selection and path planning, which works well for single objects.
However, in large-scale fruit mapping, where multiple targets must be considered, such as exploration frontiers for discovering new fruits and missing surfaces for higher coverage, it is crucial to jointly optimize both path and view selection to effectively balance mapping performance and motion cost.
Second, the learned priors in these methods are specialized for object reconstruction and may not generalize well to the highly complex and unforeseen structures in plants.
In our work, we suggest a region prior to allowing for SCP constraints in SHPP, highlighting the advantage of a unified optimization strategy for fruit mapping.

\section{System Overview} \label{S:system_overview}

\begin{figure}[!t]
\centering
\includegraphics[width=1.0\columnwidth]{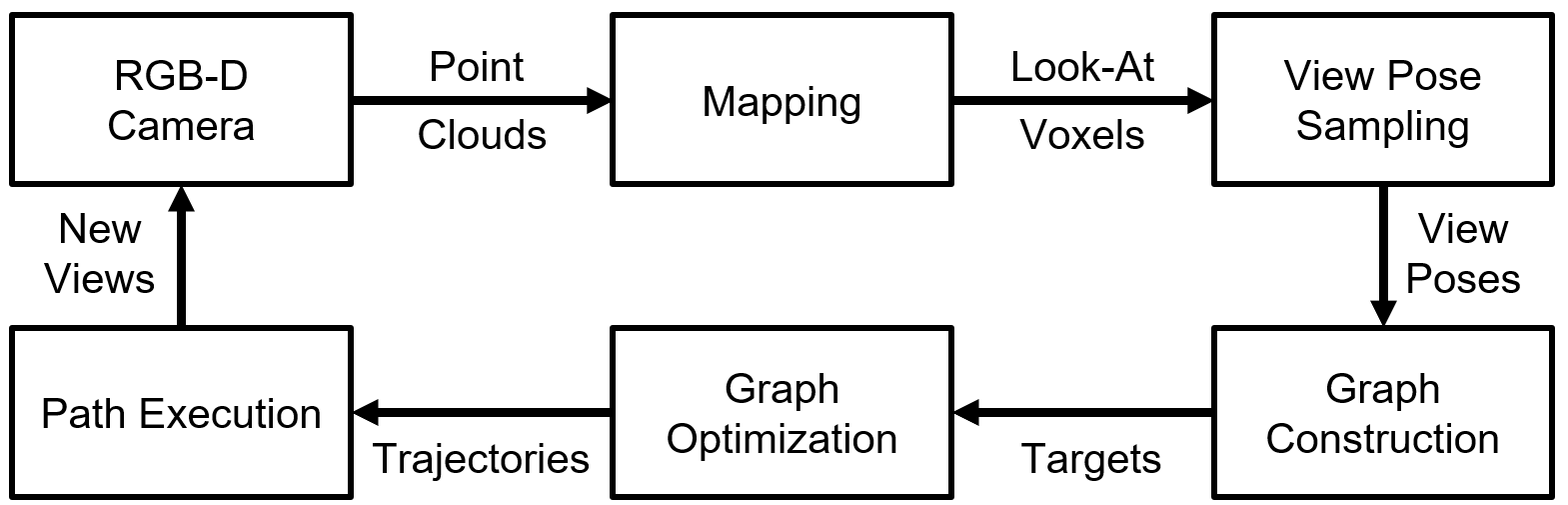}
\caption{
Overview of our system.
We use OctoMap~\citep{hornung2013ar} to fuse the camera’s point clouds into a 3D occupancy map.
From the current map state, we extract look-at voxels that guide the sampling of reachable view pose candidates.
Our planner then constructs a view motion graph and formulates a global optimization problem to determine an optimized sequence of views and connecting paths that minimizes motion costs while ensuring complete target coverage.
}
\label{fig_system}
\vspace{-0.5cm}
\end{figure}

We present a novel globally optimized view motion planning method that minimizes robot motion costs while maximizing fruit coverage in initially unknown 3D environments.
Our setup features a robot arm mounted on a trolley that moves along plant rows in a glasshouse.
The trolley is capable of vertical adjustments through lifting and lowering, allowing positioning for mapping fruits at different heights.

An overview of the proposed system is shown in Fig.~\ref{fig_system}.
We use OctoMap~\citep{hornung2013ar} with 1\,cm resolution to fuse point cloud observations from RGB-D images captured by the calibrated camera into a probabilistic 3D map consisting of free, occupied, unknown space, and ROIs, \ie, fruit voxels using the detection in point clouds~\citep{smitt2021pathobot}.
Based on the map state, we sample look-at voxels that have the potential to observe more fruit surfaces.
Afterwards, reachable view pose candidates are sampled to observe unknown space around these voxels by considering the motion constraints of the robot arm.
The planner constructs a view motion graph with full target coverage constraints, aiming to obtain the globally shortest path that minimizes motion costs.
Unlike the graph-based best-first path search in VMP~\citep{zaenker2023iros}, we formulate the global optimization to derive an optimized sequence of views and connecting paths.

In an initially unknown environment, each newly acquired view updates the map, making the existing graph outdated. 
Therefore, we update the old graph with newly sampled views and re-optimize a new sequence of views and connecting paths to incorporate the latest information.
To better balance coverage quality with motion efficiency, we schedule these re-optimizations at fixed 12-second intervals.
Similar to VMP's stop criterion, our system continues the entire cycle in Fig.~\ref{fig_system} until the allocated 60-second time budget is exhausted.
Note that it completes the final planned view paths even after the stop criterion is met.
To enhance computational efficiency, we perform mapping and trajectory execution in parallel, allowing the map to be updated during the robot movement.
We incorporate new RGB-D data only from each planned view because depth data collected while the robot is in motion tends to be too noisy and may not be relevant to our goal (\eg, looking to the ground).

\section{Globally Optimized View Motion Planning} \label{S:graph_optimization}

The key to our globally optimized view motion planning is the construction of a view motion graph and the formulation of global optimization with coverage constraints to determine the optimized sequence of views and connecting paths.

\subsection{Prior-Aware Sparse Graph Construction} \label{S:graph_construction}

The construction of our view motion graph consists of three main steps: (1) extracting look-at voxels based on the current map state and region prior, (2) sampling view poses that look at these voxels as graph vertices, and (3) sparsely connecting vertices with edges to their nearest neighbors, in which edge weights represent the robot's motion cost.

\subsubsection{Look-At Voxel Extraction}

\begin{figure}[!t]
\centering
\includegraphics[width=0.9\columnwidth]{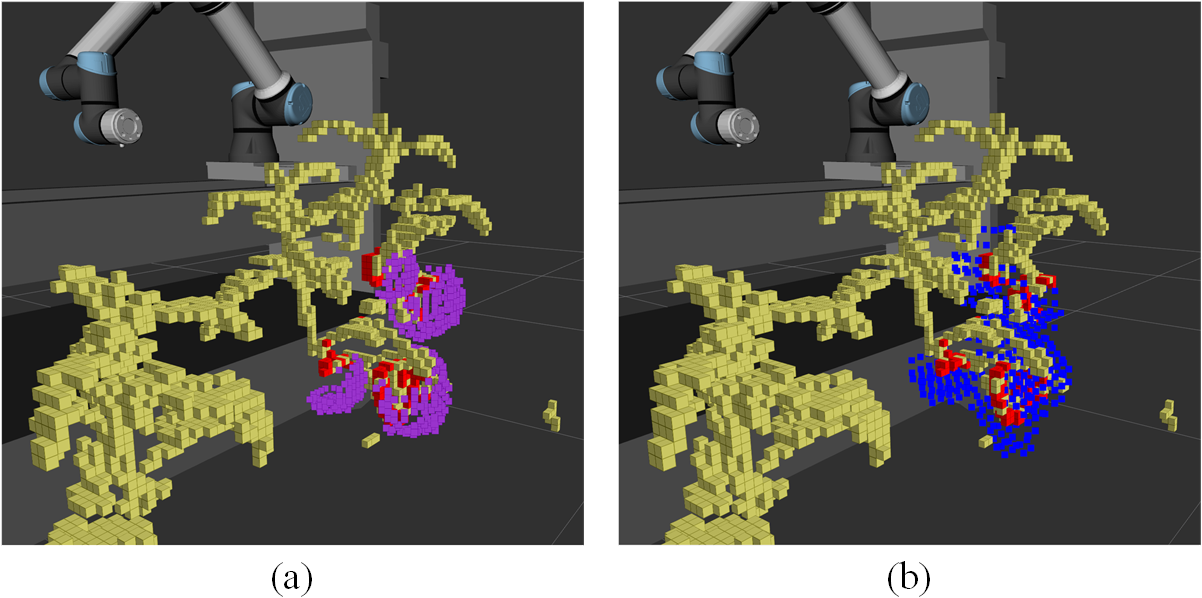}
\caption{
Examples of two types of PRIOR look-at voxels.
Occupied voxels are visualized in red for ROI fruits, while other plant parts are shown in yellow.
(a) Shape prior – Missing surface voxels derived from the occlusion-free fitted superellipsoid~\cite{menon2023iros}, visualized in purple.
(b) Region prior – Unknown voxels extracted from an inflated octree, which expands ROI regions, visualized in blue.
}
\label{fig_priors}
\vspace{-0.5cm}
\end{figure}

Following VMP~\citep{zaenker2023iros}, we identify informative look-at voxels that are likely to contain fruits in their surrounding region.
VMP employs three types of look-at voxels: OCC-UNK, FRE-UNK, and ROI-UNK.
OCC-UNK represents the occupied-unknown boundary, which we use because it helps detect hidden fruits occluded by other plant parts.
FRE-UNK represents the free-unknown boundary, which we use because it helps detect additional plants.
ROI-UNK represents the ROI-unknown boundary, which we use because it helps complete the fruit surface.
In this work, we introduce an additional type of look-at voxels called PRIOR.
We consider two types of priors: (1) shape prior, which is derived from the occlusion-free superellipsoid fitting method~\cite{menon2023iros}, and (2) region prior, which is obtained using an inflated octree with a resolution set to half of the OctoMap resolution to expand ROI regions within a distance of 10\,cm.
Fig.~\ref{fig_priors} illustrates both types of priors. 
To enhance computational efficiency, PRIOR voxels are extracted in parallel.
To balance exploration and exploitation, we randomly sample the look-at voxels with the following probabilities: 30\% OCC-UNK, 20\% FRE-UNK, 35\% ROI-UNK, and 15\% PRIOR.

\subsubsection{View Pose Sampling}

\begin{figure}[!t]
\centering
\includegraphics[width=0.65\columnwidth]{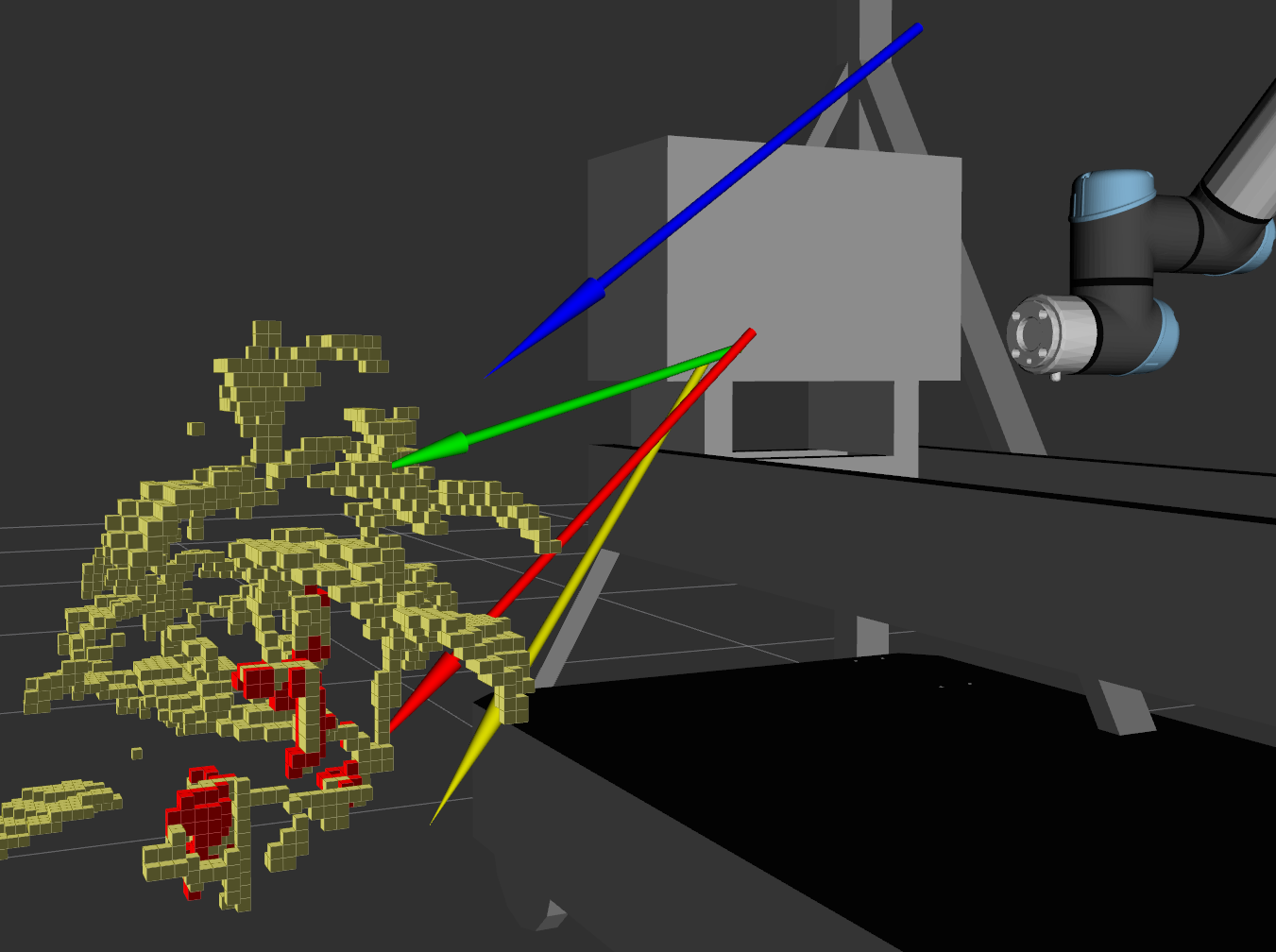}
\caption{
Example of view pose sampling under four types of look-at voxels.
Occupied voxels are visualized in red for ROI fruits, while other plant parts are shown in yellow.
The arrows represent the camera's viewing direction toward the corresponding look-at voxel for which the view was sampled.
The arrow color indicates the type of look-at voxel: Red for ROI-UNK, Green for OCC-UNK, Blue for FRE-UNK, and Yellow for PRIOR (inflated unknown voxels of the region prior in this example). 
}
\label{fig_viewpose}
\vspace{-0.5cm}
\end{figure}

A view pose is defined by the 3D position and orientation in $R^3\times SO(3)$ space.
We sample a set of view poses whose orientation is pointing to these look-at voxels.
First, the 3D position is randomly sampled within the arm’s workspace and the sensor’s operational range.
Next, we use ray-casting in OctOMap to ensure a clear line of sight from the view position to the look-at voxel.
Finally, inverse kinematics is used to validate the existence of a collision-free joint configuration of the view pose.
A view pose is accepted only if all conditions are met.
Fig.~\ref{fig_viewpose} illustrates sampled view poses for four types of look-at voxels.
To enhance computational efficiency, view pose sampling is performed in parallel.

These sampled view poses serve as potential vertices in our view motion graph.
Inspired by the view similarity method~\citep{menon2023iros}, each newly generated pose is compared to existing vertices based on viewing angles (cosine distance) and motion proximity that is assessed using joint-space distance instead of the original Euclidean distance.
Poses deemed similar are excluded to avoid redundancy, while those that pass the similarity check are added to the graph.

\subsubsection{Motion-Aware Edge Connection}

We establish edges by connecting each vertex to its neighboring vertices, forming an undirected view motion graph.
To maintain sparsity for efficient global optimization, each vertex is linked to its five nearest neighbors.
The distance between vertices is determined by the trajectory distance in robot joint space, calculated based on the difference in their joint configurations using MoveIt~\citep{chitta2016ros}.
Although this distance does not explicitly model physical constraints such as joint-specific torque or energy consumption, it serves as a practical approximation that effectively reflects the motion cost of the manipulator.
Each edge weight corresponds to this trajectory distance, representing the motion cost associated with transitioning between two view poses.
An edge is created only if the intermediate poses along the trajectory are valid and collision-free, ensuring feasible motion execution. 

\subsection{Global Optimization on Graph} \label{S:global_optimization}

Our key idea is to formulate a global optimization that simultaneously minimizes robot motion costs while maximizing fruit coverage.
To this end, we adopt the SHPP to compute the globally shortest path while integrating SCP constraints to ensure full coverage of the selected targets.
The original SHPP aims to find a sequence of views and connecting paths that minimizes the total path length while ensuring that each view on the graph is visited exactly once.
Starting from a given view vertex (in our case, the robot’s current position), SHPP determines a path that terminates at another view vertex.

However, visiting all views on the graph exactly once is not desirable for our application.
Due to the large sensor field of view, many views have overlapping regions, making it inefficient to include each view in the path.
Therefore, we incorporate SCP constraints to select a subset of views while ensuring full coverage of selected targets. 
The SCP constraints enforce that each target voxel is covered by at least one view that can observe it.
Given the NP-hardness of the problem and the need to improve computational efficiency, we reduce the number of optimization constraints by selecting a subset of look-at voxels as optimization targets.

\subsubsection{Coverage Target Selection}

Our goal is to prioritize the coverage of existing fruit regions, so we incorporate only ROI-UNK and PRIOR look-at voxels into our SCP constraints.
Although OCC-UNK and FRE-UNK look-at voxels are not included as coverage targets, our system will select views for exploration to discover additional fruits because half of the sampled views in the graph are generated to observe these voxels.

\subsubsection{Integer Linear Programming Formulation} 

We define the following integer linear programming (ILP) formulation for our global optimization:
\begin{equation*}
\label{equ:ILP}
\resizebox{1.0\columnwidth}{!}{ 
$
\begin{aligned}
\min: \, & \sum_{i=0}^{n} \sum_{j=0}^{n} m_{ij} p_{ij},  && \forall i \neq j, \\
\mathrm{s.t.}: \,
& (a) \, p_{ij} \in \{0,1\},  && \forall i, j \in \mathbb{N}_0^n, \\
& (b) \, \sum_{j=0}^{n} p_{ji} \leq 1, \, \sum_{j=0}^{n} p_{ij} \leq 1, \, \sum_{j=0}^{n} p_{ji} = \sum_{j=0}^{n} p_{ij}, && \forall i \in \mathbb{N}_0^n, \\
& (c) \, \sum_{i,j \in S} p_{ij} \leq |S|-1,  && \forall S \subset \mathbb{N}_0^n, \\
& (d) \, v_i \in \{0,1\},  && \forall i \in\mathbb{N}_0^n, \\
& (e) \, \sum_{i=0}^{n} v_i r_{ic} \geq 1, && \forall c \in C, \\
& (f) \, \sum_{j=0}^{n} p_{ji} \leq v_i, \, \sum_{j=0}^{n} p_{ij} \leq v_i, \, \sum_{j=0}^{n} (p_{ij} + p_{ji}) \geq v_i, && \forall i \in \mathbb{N}_0^n, \\
\end{aligned}
$
}
\end{equation*}

We transform the SHPP into the traveling salesman problem (TSP), the variant where the path starts from a given view vertex and returns to the same vertex to form a tour.
Given the set $\{0, \ldots, n-1\}$ representing the indexes of view vertices in our graph, we introduce a virtual end vertex with index $n$.
This virtual vertex is connected to all view nodes using zero-weighted edges, ensuring that the tour can properly return to the starting viewpoint.
Thus, the extended set $\mathbb{N}_0^n=\{0, \ldots, n\}$ represents all views.
This transformation allows us to define the ILP above.

\begin{figure}[!t]
\centering
\includegraphics[width=0.7\columnwidth]{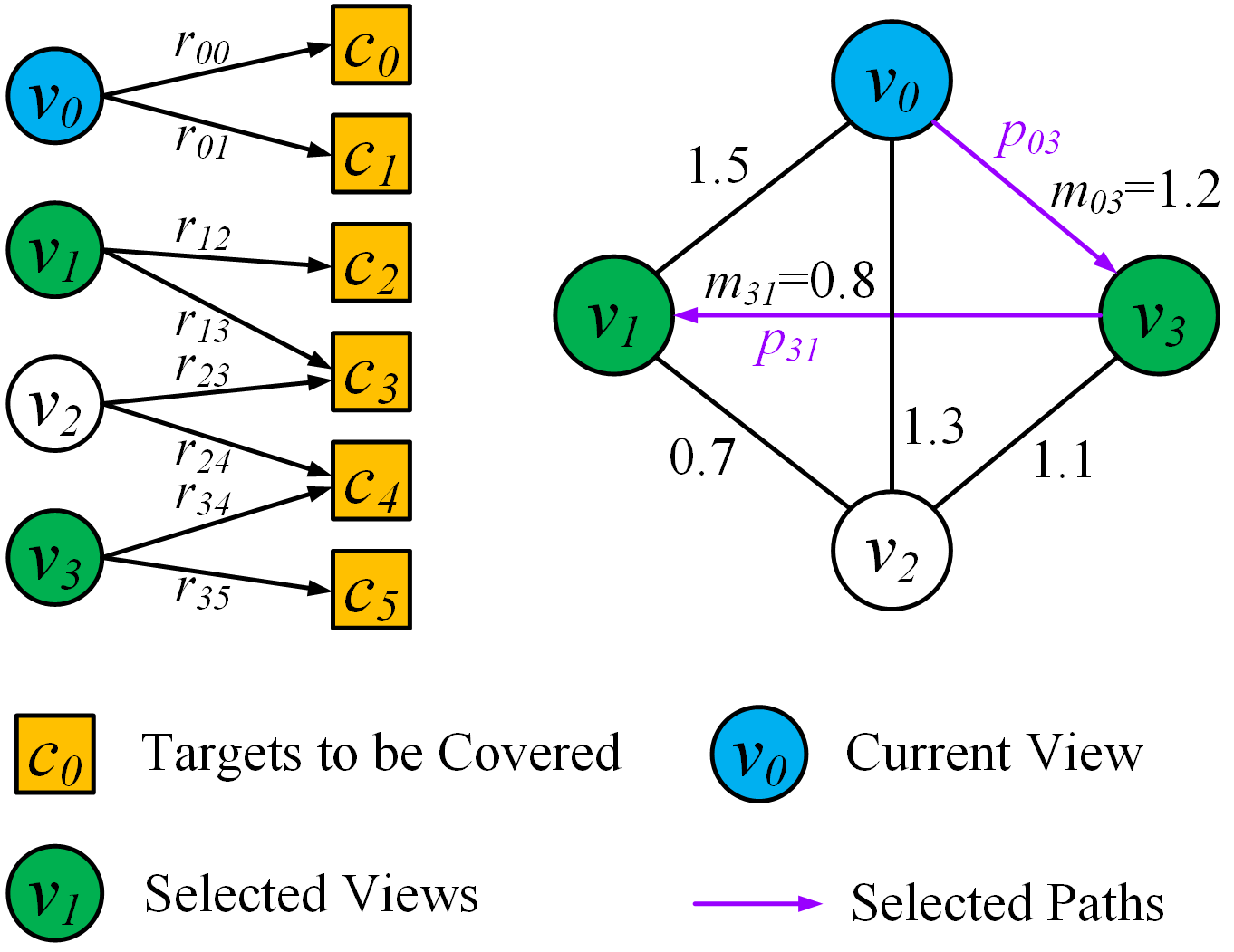}
\caption{
An example of our ILP solution under four sampled views.
\textbf{Left:} $c_i$ represents the target voxels that need to be covered, while $r_{ij}$ indicates the visibility from different views, determined through ray-casting.
\textbf{Right:} The generated view motion graph, where each edge is associated with a motion cost $m_{ij}$.
The output from the optimizer consists of the newly selected views ($v_1$ and $v_3$) along with the selected paths ($p_{03}$ and $p_{31}$) connecting them from the current view $v_0$.
For a simplified illustration, we omit the virtual end vertex and display only relevant constants $r_{ij}=1$ and variables $p_{ij}=1$.
The solution is optimized to achieve (1) full coverage of all six target voxels ($c_0$ to $c_5$), and (2) minimal motion cost of 2.0 ($m_{03}$ plus $m_{31}$).
}
\label{fig_ILP}
\vspace{-0.5cm}
\end{figure}

The objective function $\min\sum_{i=0}^{n} \sum_{j=0}^{n} m_{ij} p_{ij}$ is to minimize the total motion costs associated with the paths selected.
Constants $m_{ij} \in \mathbb{R}$ are the motion cost from edge weights in our graph.
First, it is subject to path constraints:
\begin{itemize}
    \item[(a)] $p_{ij}$ are binary decision variables for path selection, indicating whether a path between two view vertices $i$ and $j$ is selected or not.
    \item[(b)] These constraints ensure that each view is visited at most once from path connections. For each view $i$, $\sum_{j=0}^{n} p_{ji} \leq 1$ ensures that at most one incoming path connection leads to each view vertex; $\sum_{j=0}^{n} p_{ij} \leq 1$ ensures that at most one outgoing path connection leads to each view vertex; and $\sum_{j=0}^{n} p_{ji} = \sum_{j=0}^{n} p_{ij}$ enforces that if a view vertex has an incoming path connection, it must also have a corresponding outgoing path connection, maintaining path continuity.
    \item[(c)] These constraints are designed to eliminate subtours, as solutions containing multiple disjoint cycles would be feasible. A subtour is a cycle that does not visit all the selected vertices. To prevent subtours, for every proper subset of vertices where \mbox{$S: |S| \geq 2$}, $\sum_{i,j \in S} p_{ij} \leq |S|-1$ ensure that the number of selected paths in $S$ does not equal or exceed the number of vertices in $S$. 
\end{itemize}

Second, the foundation for defining the SCP constraints in view planning determines which voxels can be seen from a given view.
We employ ray-casting in OctoMap to verify whether a target voxel $c \in C$ is observable from a view $i$, where $C$ is the set of selected target voxels.
The results are stored as binary constants $r_{ic}$, where $r_{ic}=1$ indicates that voxel $c$ is observed, and $r_{ic}=0$ otherwise.
Thus, the problem is subject to coverage constraints:
\begin{itemize}
    \item[(d)] $v_i$ are binary decision variables for view selection, indicating whether a view $i$ is selected or not.
    \item[(e)] These constraints ensure that all selected targets are covered at least once. $\sum_{i=0}^{n} v_i r_{ic} \geq 1$ ensures that if a target voxel $c$ is selected, at least one view capable of observing it must also be selected.
\end{itemize}

Third, since two different types of decision variables are correlated, the problem is subject to relation constraints:
\begin{itemize}
    \item[(f)] $\sum_{j=0}^{n} p_{ji} \leq v_i$ enforces that ensures that if view $i$ is not selected, then no incoming path connection to the view should exist, and if view $i$ is selected, then at most one incoming path connection to the view is allowed; $\sum_{j=0}^{n} p_{ij} \leq v_i$ enforces that if view $i$ is not selected, then no outgoing path connection to the view should exist, and if view $i$ is selected, then at most one outgoing path connection to the view is allowed; $\sum_{j=0}^{n} (p_{ij} + p_{ji}) \geq v_i$ enforces that if view $i$ is selected, then at least one incoming or outgoing path connection (or both) must exist to or from the view.
\end{itemize}

Finally, in our setup, the view $0$ is always designated as the start view vertex, representing the robot’s current position, while the view $n$ serves as the virtual end vertex.
Thus, specific decision variables for these two views must be predefined as:
\begin{itemize}
\item $v_0 = v_n = 1$ ensures that both the start and virtual end views are always selected.
\item $p_{0n} = 0, p_{n0} = 1$ enforces that the virtual end view is always connected directly back to the start view, maintaining the required tour.
\end{itemize}

We employ the Gurobi optimizer, a linear programming solver~\citep{gurobi2021gurobi}, to compute the solution for the problem.
We present an instance solution in Fig.~\ref{fig_ILP}.
To prevent infinite optimization time due to the NP-hardness, we impose a maximum optimization time limit of 20 seconds.

\section{Experiments} \label{S:experimental_results}

Our experiments are designed to show that (1) a sparse graph structure combined with a region-prior-based target selection strategy yields the effective sequence of views and connecting paths when solving our global optimization within a limited planning time, (2) compared to existing view planning baselines, our globally optimized view motion planner (GO-VMP) achieves a superior balance between fruit mapping performance and robot motion cost, and (3) our system is viable for deployment in real-world environments. 
For more details, please refer to the accompanying video at \url{https://youtu.be/mSWlDTn5ykg}.

\subsection{Setup and Evaluation}

\begin{figure}[!t]
\centering
\includegraphics[width=0.96\columnwidth]{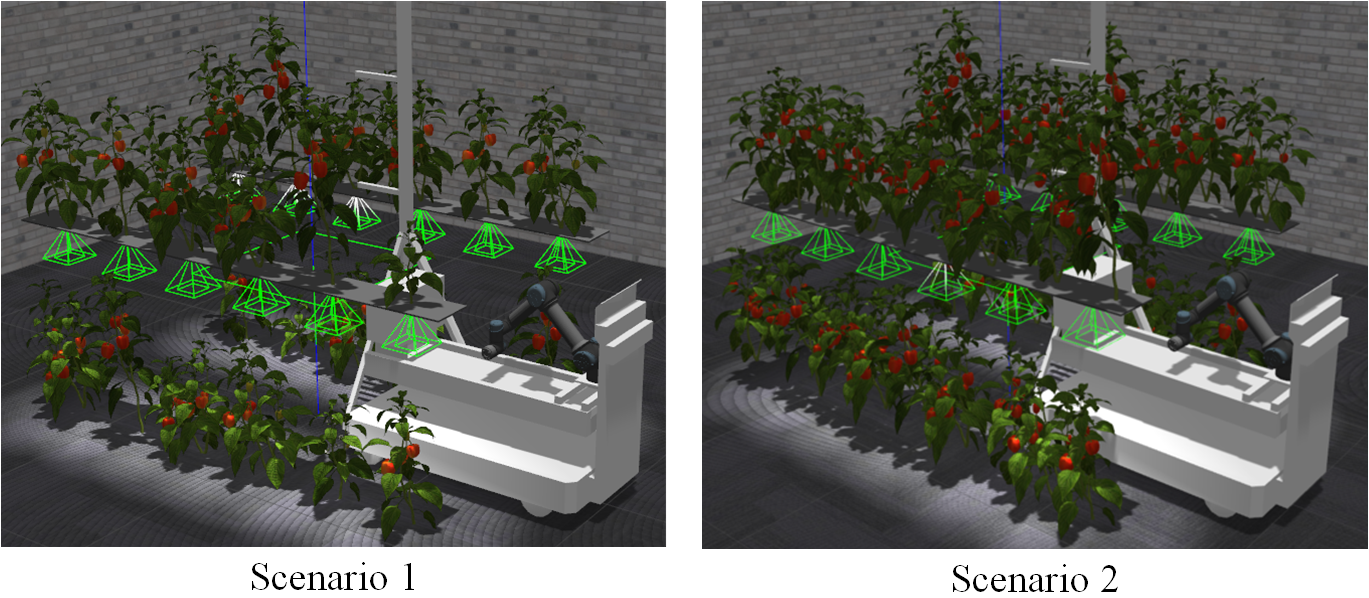}
\caption{
The glasshouse simulation consists of two scenarios with plants arranged in two rows, with each row containing plants positioned at two distinct height levels.
In Scenario 1, there are 24 plants with 120 sweet pepper fruits, while Scenario 2 contains 36 plants with 217 fruits and has more occlusions. 
A robot arm equipped with an RGB-D camera is mounted on a trolley platform that moves between the two rows and vertically adjusts the robot arm by lifting and lowering it.
}
\label{fig_simulation}
\vspace{-0.5cm}
\end{figure}

We designed two glasshouse simulation scenarios in Gazebo~\citep{koenig2004design} that closely mimic real-world deployment conditions, as shown in Fig.~\ref{fig_simulation}. 
We exploit the distinctive red coloration of sweet peppers by performing a Hue-Saturation-Intensity value comparison for easy fruit detection.
Realistic point clouds are simulated using an RGB-D camera on the arm’s end effector, incorporating Gaussian noise.
Due to the robot's limited workspace, we follow VMP~\citep{zaenker2023iros} to partition the entire crop row into 16 equally sized segments.
The overall mission is to map all fruits in each segment under a limited time budget, with the robot sequentially performing the mapping process in each segment.
Simulation experiments are run on a computer with an i7-13700H CPU, 16 GB RAM, and RTX4060 GPU with ROS Noetic.

For a fair comparison, all planners operate under the same setup as our system, with a time budget of 60 seconds per segment as a stop criterion.
To evaluate fruit mapping performance, fruit clusters are additionally detected using our previously developed superellipsoid fitting method~\citep{marangoz2022case}.
The number of \textbf{detected fruits} is determined by counting clusters that successfully match a corresponding ground truth cluster, based on a center-to-center distance threshold of 20\,cm.
Fruit \textbf{surface coverage} is computed as the ratio of matched ground truth voxels on the detected superellipsoid to the total number of voxels in the superellipsoid.
To assess the accuracy of volume estimation, we fit a 3D convex hull to the detected fruit surfaces and compute \textbf{volume accuracy} as the ratio of the convex hull volume to the ground truth volume.
These three mapping metrics are reported as the average values computed across all ground truth fruits.
The robot's \textbf{motion cost} is measured as the total accumulated joint space distance between joint configurations of view poses.

\subsection{Target Selection and Graph Sparsity} \label{S:Ablation}

This section examines the impact of different optimization targets and graph sparsity on global optimization under a limited planning time (20 seconds discussed in Sec.~\ref{S:global_optimization}).
We consider five different types of targets that are included in the coverage constraint as detailed in Sec.~\ref{S:graph_construction}:
\begin{itemize}
    \item ROI-UNK: Only ROI-UNK voxels.
    \item SE: Only PRIOR voxels from the superellipsoid.
    \item Infl: Only PRIOR voxels from the Inflated octree.
    \item SE+ROI-UNK: A combination of SE and ROI-UNK.
    \item Infl+ROI-UNK: A combination of Infl and ROI-UNK.
\end{itemize}
Table~\ref{tab_targets} reports our planner's performance using different types of targets.
The results indicate that: (1) Combining ROI-UNK targets from both SE and Infl priors is essential.
This is expected, as exploitation frontiers are likely to observe more new fruit surfaces.
(2) The region prior Infl, outperforms the shape prior SE.
This is reasonable because the superellipsoid serves only as an approximation of the unknown fruit surface and may introduce some inaccuracies.
Additionally, considering the entire region of unknown voxels around the fruits has the potential to observe more unknown surfaces.

\begin{table}[!t]
\centering
\resizebox{\columnwidth}{!}{%
\begin{tabular}{ccccc}
\hline
\makecell{Target\\Type} & \makecell{Detected\\Fruits} & \makecell{Surface\\Coverage\,(\%)} & \makecell{Volume\\Accuracy\,(\%)} & \makecell{Motion\\Cost} \\
\hline
ROI-UNK                 & \textbf{106.7} ± 4.9 & 63.0 ± 2.4 & 66.4 ± 1.5 & 202.5 ± 5.7 \\
SE                  & 102.7 ± 4.5 & 59.7 ± 1.2 & 65.3 ± 1.9 & 210.5 ± 6.8 \\
Infl                & 101.5 ± 2.5 & 62.1 ± 1.4 & 65.2 ± 2.3 & 207.8 ± 7.3 \\
SE+ROI-UNK              & 104.5 ± 4.7 & 61.9 ± 2.5 & \textbf{67.6} ± 0.6  & 222.3 ± 8.9 \\
Infl+ROI-UNK\,(Ours)    & 106.6 ± 1.4 & \textbf{64.2} ± 0.2 & 67.5 ± 1.0 & \textbf{190.0} ± 10.4 \\
\hline
\end{tabular}%
}
\caption{
Ablative study on types of targets in the coverage constraint.
Each method is executed three times using our planner under Scenario 1, and the mean and standard deviation of metrics are reported.
As can be seen, the Infl+ROI-UNK method achieves higher or similar mapping performance while reducing motion cost.
}
\label{tab_targets}
\vspace{-0.2cm}
\end{table}

\begin{table}[!t]
\centering
\resizebox{\columnwidth}{!}{%
\begin{tabular}{ccccc}
\hline
\makecell{Graph\\Sparsity} & \makecell{Detected\\Fruits} & \makecell{Surface\\Coverage\,(\%)} & \makecell{Volume\\Accuracy\,(\%)} & \makecell{Motion\\Cost} \\
\hline
Com             & 103.0 ± 1.7 & 63.7 ± 0.7 & \textbf{69.0} ± 2.6 & 402.4 ± 52.7 \\
Den             & 100.7 ± 4.7 & 61.9 ± 0.9 & 67.5 ± 2.3  & 210.0 ± 2.3 \\
Spa\,(Ours)     & \textbf{106.6} ± 1.4 & \textbf{64.2} ± 0.2 & 67.5 ± 1.0 & \textbf{190.0} ± 10.4 \\
\hline
\end{tabular}%
}
\caption{
Ablative study on different graph sparsity levels.
Each method is executed three times using our planner under Scenario~1, and the mean and standard deviation of metrics are reported.
As can be seen, the sparse graph reduces motion cost while maintaining higher or comparable mapping performance.
}
\label{tab_sparsity}
\vspace{-0.5cm}
\end{table}

We consider three different levels of graph sparsity:
\begin{itemize}
    \item Com (Complete Graph): Fully connected views with all possible edges.
    \item Den (Dense Graph): Each view is connected to its 10 nearest neighbors.
    \item Spa: Spa (Sparse Graph): Each view is connected to its 5 nearest neighbors.
\end{itemize}
Table~\ref{tab_sparsity} reports our planner's performance under different sparsity levels.
The results indicate that the proposed sparse graph is sufficient to achieve the most effective sequence of views and connecting paths within a limited optimization time. 
Note that we evaluate our two strategies on Scenario~1, and then apply them to Scenario 2 for novel testing against baselines in the next section, which highlights their generalizability to more complex occlusion conditions.

\subsection{Evaluation of Fruit Mapping}

\begin{figure*}[!t]
\centering
\includegraphics[width=0.9\textwidth]{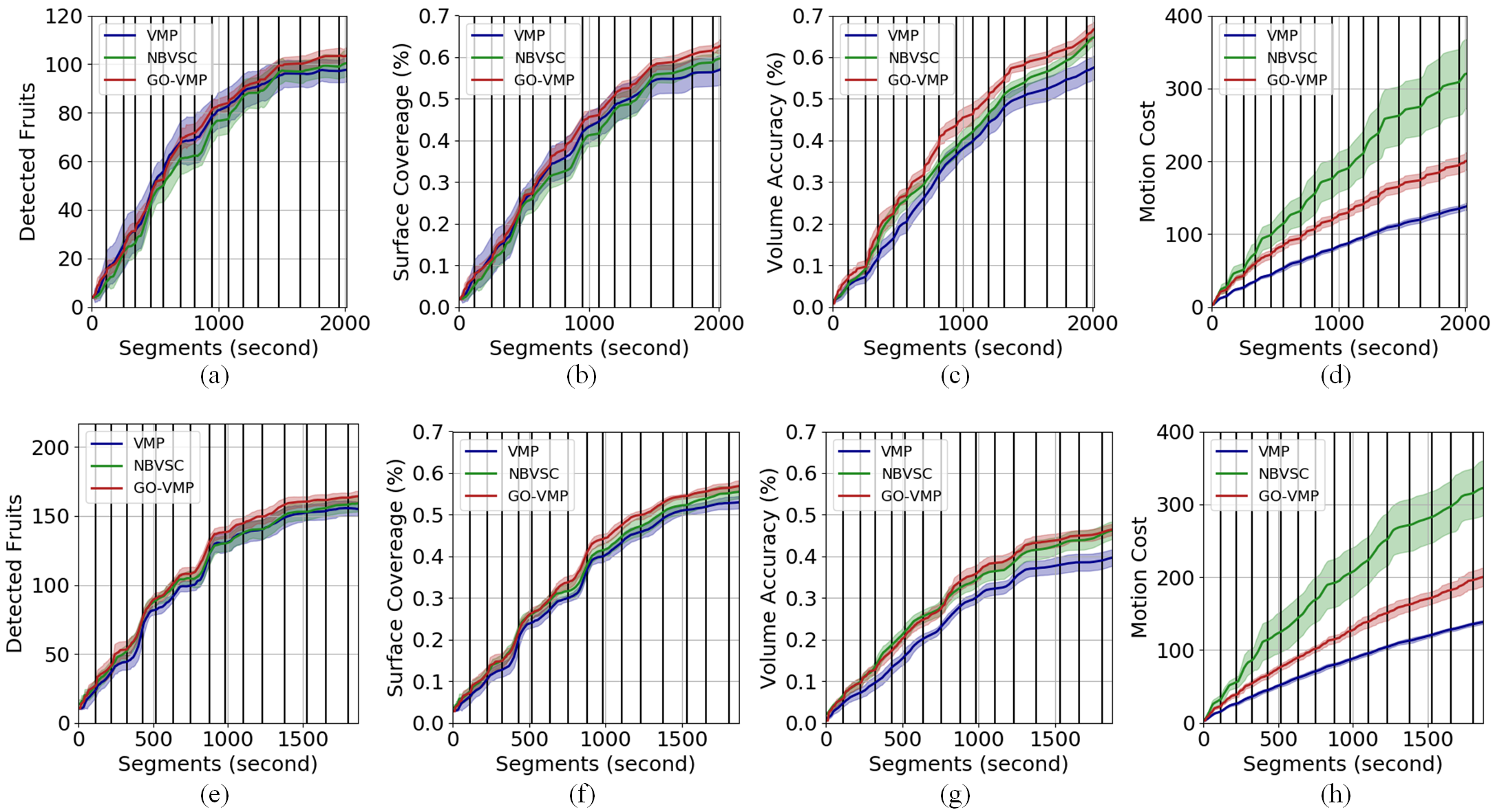}
\caption{
Evaluation of our planner’s performance in fruit mapping over time, compared to VMP~\citep{zaenker2023iros} and NBV-SC~\citep{menon2023iros}, across Scenario 1 (a–d) and Scenario 2 (e–h).
Each method is executed 10 times, and the mean and standard deviation of metrics are reported.
To account for variations in execution time across different trials and ensure a smooth and comparable visualization, we apply linear interpolation within each segment and then stack the interpolated segments together to form the final curve.
The bold vertical black lines indicate segment boundaries, facilitating a clearer analysis of per-segment performance.
As can be seen, in both scenarios, the proposed GO-VMP detects more fruits, enhances surface coverage, and achieves higher volume accuracy than VMP with a moderate increase in motion cost, while significantly reducing motion cost compared to NBV-SC over all the segments.
}
\label{fig_baselines}
\vspace{-0.2cm}
\end{figure*}

\begin{table*}[!t]
\centering
\resizebox{0.8\textwidth}{!}{%
\begin{tabular}{cccccc}
\hline
Scenario & Planner & Detected Fruits & Surface Coverage\,(\%) & Volume Accuracy\,(\%) & Motion Cost \\
\hline
\multirow{3}{*}{1} & VMP              & $^{\star}$97.7 ± 5.2  & $^{\star\star\star}$57.1 ± 3.7 & $^{\star\star\star}$57.6 ± 2.9 & \textbf{138.3} ± 4.9  \\
                   & NBV-SC           &  100.4 ± 5.7 & $^{\star\star}$59.7 ± 2.3 & $^\star$64.8 ± 2.1  & 320.2 ± 49.8 \\
                   & GO-VMP (Ours)    & \textbf{103.6} ± 3.8 & \textbf{62.8} ± 1.7 & \textbf{66.8} ± 1.8  & 200.3 ± 12.6 \\
\hline
\multirow{3}{*}{2} & VMP              & $^{\star\star\star}$155.2 ± 5.3 &  $^{\star\star\star}$53.0 ± 1.6 & $^{\star\star\star}$39.7 ± 2.1 & \textbf{138.3} ± 2.7  \\
                   & NBV-SC           & $^\star$159.2 ± 5.9 & 55.5 ± 1.6 & 46.2 ± 2.3 & 322.2 ± 39.9 \\
                   & GO-VMP (Ours)    & \textbf{164.3} ± 4.1 & \textbf{56.9} ± 1.4 & \textbf{46.4} ± 1.4 & 200.2 ± 13.4 \\
\hline
\end{tabular}%
}
\caption{
Evaluation of our planner's performance in comparison to VMP~\citep{zaenker2023iros} and NBV-SC~\citep{menon2023iros}.
Each method is executed 10~times and the mean and standard deviation of metrics are reported.
As all planners utilize stochastic view pose sampling, we apply Welch's \textit{t}-test at a significance level of 0.05.
$\star$, $\star\star$, $\star\star\star$ denote statistically significant results for GO-VMP compared to each baseline with \textit{p}-values thresholds of $0.05$, $0.01$, $0.001$ respectively.
These results further validate our claim that GO-VMP achieves a superior balance between fruit mapping performance and robot motion cost.
}
\label{tab_baselines}
\vspace{-0.5cm}
\end{table*}

Next, we compare our planner's performance against the motion-efficient VMP~\citep{zaenker2023iros} and the coverage-focused NBV-SC method~\citep{menon2023iros}.
The performance over time is illustrated in Fig.~\ref{fig_baselines}, while the final performance metrics are summarized in Table~\ref{tab_baselines}.
The results indicate that: (1) Compared to the motion-efficient VMP, our \mbox{GO-VMP} achieves significantly higher fruit detection, improved surface coverage, and enhanced volume accuracy, with a moderate increase in motion cost.
(2) Compared to the coverage-focused NBV-SC, our GO-VMP significantly reduces motion cost while delivering slightly better mapping performance, although not all improvements reach statistical significance.
(3) In Scenario 2, all planners exhibit a decrease in performance, which is expected due to increased occlusions from more plants and fruits. 
Nevertheless, GO-VMP follows a similar performance trend as observed in Scenario 1.
All these results confirm that our proposed global optimization for view motion planning achieves a superior balance between fruit mapping performance and robot motion cost. 
This balance is particularly valuable because high-quality fruit mapping is crucial for downstream tasks such as harvesting, while reduced robot motion cost can save electrical resources and lower operational risks.

\subsection{Timing Breakdown and Analysis}

To offer a detailed perspective on how our method balances fruit mapping performance and system efficiency, we categorize the total mission time into two primary components: (1) planning time, and (2) mapping and execution time (combined because mapping is updated in parallel during execution).
The rest of the mission time is the other supporting time, such as the trolley's travel between segments and the robot arm's movement to its home position.
Table~\ref{tab_times} shows that, under a similar overall runtime, our method reduces execution time through multi-step path planning and shortens mapping time by updating the map in parallel during execution.
These improvements free up more computational resources for planning, enabling the selection of more informative views.
In contrast, NBV-SC experiences higher execution time, while VMP allocates more time to mapping, highlighting the advantages of our design.

\subsection{Real-World Experiments}

\begin{figure}[!t]
\centering
\includegraphics[width=0.54\columnwidth]{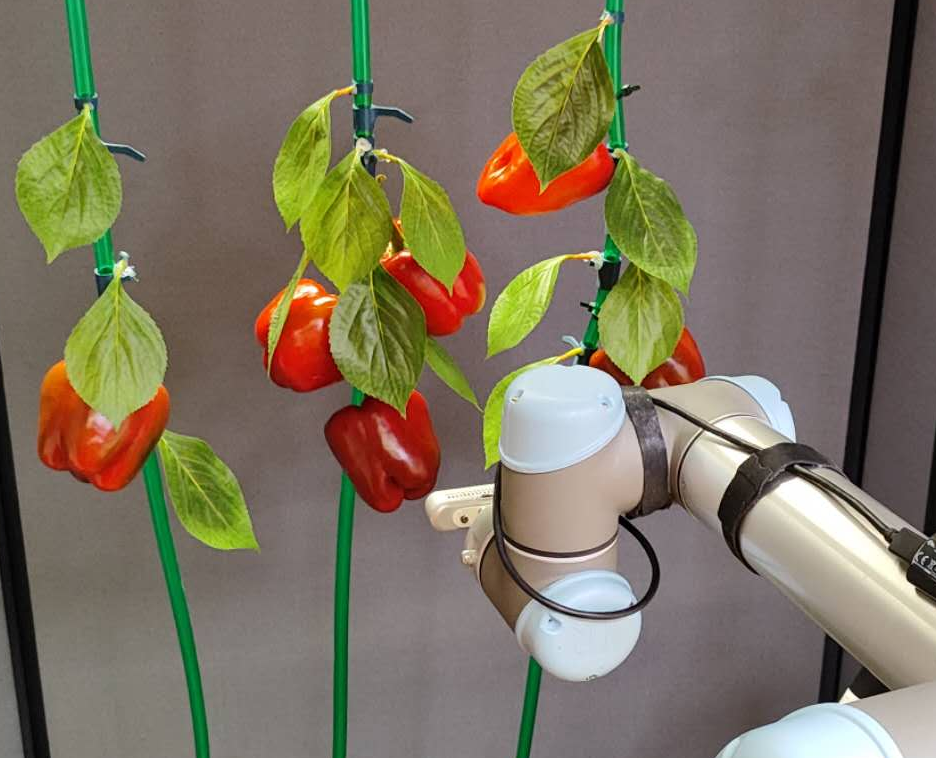}
\caption{
The real-world setup of our indoor mock-up consists of real sweet peppers, artificial leaves, and thin pipes representing stems.
}
\label{fig_realworld}
\vspace{-0.5cm}
\end{figure}

We deployed our approach in a real-world environment using a UR5 robot arm with an Intel Realsense D435 camera mounted on its end-effector.
Due to the off-season in the commercial glasshouse, we constructed an indoor mock-up of sweet pepper plants as shown in Fig.~\ref{fig_realworld} and conducted a mapping test using real data to demonstrate the practical applicability of our GO-VMP system.
The mapping process and results are showcased in the accompanying video.

\begin{table}[!t]
\centering
\resizebox{1.0\columnwidth}{!}{%
\begin{tabular}{ccccc}
\hline
Planner & Runtime\,(s) & Planning\,(s) & \makecell{Mapping\,\&\\Execution\,(s)} & \makecell{View\\Number}\\
\hline
VMP            & 1,914 ± 56 & 313 ± 7  & 1,016 ± 46 & 68.8 ± 3.0 \\
NBV-SC         & 1,716 ± 41 & \textbf{102} ± 9  & 1,006 ± 27 & 57.6 ± 2.9 \\
GO-VMP (Ours)  & \textbf{1,704} ± 48 & 659 ± 35 & \textbf{618} ± 22  & 101.4 ± 3.8 \\
\hline
\end{tabular}%
}
\caption{
Time breakdown of the entire mission and the number of views executed in comparison to baselines.
The mean and standard deviation of the reported metrics are computed by averaging over all 20 trials across both scenarios.
As can be seen, our method reduces mapping and execution time while allocating more time for planning to identify more informative views.
}
\label{tab_times}
\vspace{-0.2cm}
\end{table}

\section{Conclusions} \label{S:conclusions}

In this work, we presented an approach to globally optimized view motion planning for fruit mapping. 
We formulate the global optimization by integrating the set covering problem with the shortest Hamiltonian path problem, creating a unified framework that computes a global view path with minimized robot motion, while ensuring full coverage of the selected targets.
To mitigate the NP-hard nature of this global optimization in fruit mapping, our planner builds a sparse graph and employs a region-prior-based target selection strategy, thereby achieving effective outcomes within a limited planning time.
The experimental results demonstrate that our method achieves a superior balance between fruit mapping performance and robot motion cost compared to both motion-efficient and coverage-focused baselines. 

\bibliographystyle{IEEEtranSN}
\footnotesize
\bibliography{iros2025}

\end{document}